\documentclass[10pt,twocolumn,letterpaper]{article}

\usepackage[pagenumbers]{cvpr} 

\usepackage{amsmath, amssymb, amsthm}
\usepackage{graphicx}
\usepackage{xspace}
\usepackage{subcaption}
\usepackage{multirow}
\usepackage{soul}
\usepackage{pifont}
\usepackage{colortbl}
\usepackage{afterpage}

\definecolor{cvprblue}{rgb}{0.21,0.49,0.74}
\usepackage[pagebackref,breaklinks,colorlinks,allcolors=cvprblue]{hyperref}

\def\paperID{3746} 
\def\confName{CVPR}
\def\confYear{2025}

\title{Humans as Checkerboards: Calibrating Camera Motion Scale \\
for World-Coordinate Human Mesh Recovery}

\author{Fengyuan Yang, Kerui Gu, Ha Linh Nguyen, Tze Ho Elden Tse, Angela Yao\\
  National University of Singapore\\
  {\tt\small \{fyang, keruigu, hlinhn, eldentse, ayao\}@comp.nus.edu.sg}
}

\begin{document}

\maketitle
\begin{abstract}

Accurate camera motion estimation is essential for recovering global human motion in world coordinates from RGB video inputs. 
SLAM is widely used for estimating camera trajectory and point cloud, but monocular SLAM does so only up to an unknown scale factor. 
Previous works estimate the scale factor through optimization, but this is unreliable and time-consuming.
This paper presents an optimization-free scale calibration framework, \textbf{H}uman \textbf{a}s \textbf{C}heckerboard (\textbf{HAC}).  HAC innovatively leverages the human body predicted by human mesh recovery model as a calibration reference.  Specifically, it uses the absolute depth of human-scene contact joints 
as references to calibrate the corresponding relative scene depth from SLAM. 
HAC benefits from geometric priors encoded in human mesh recovery models to estimate 
the SLAM scale and achieves precise global human motion estimation.
Simple yet powerful, our method sets a new state-of-the-art performance for global human mesh estimation tasks, reducing motion errors by 50\% over prior local-to-global methods while using 100$\times$ less inference time than optimization-based methods. 
See: \url{https://martayang.github.io/HAC/}.
\vspace{-0.4cm}
\end{abstract}    
\section{Introduction}
\label{sec:intro}

\begin{figure}[t!]
    \vspace{-0.6cm}
    \centering
    \begin{subfigure}[b]{0.5\textwidth}
        \centering
        \includegraphics[width=\textwidth]{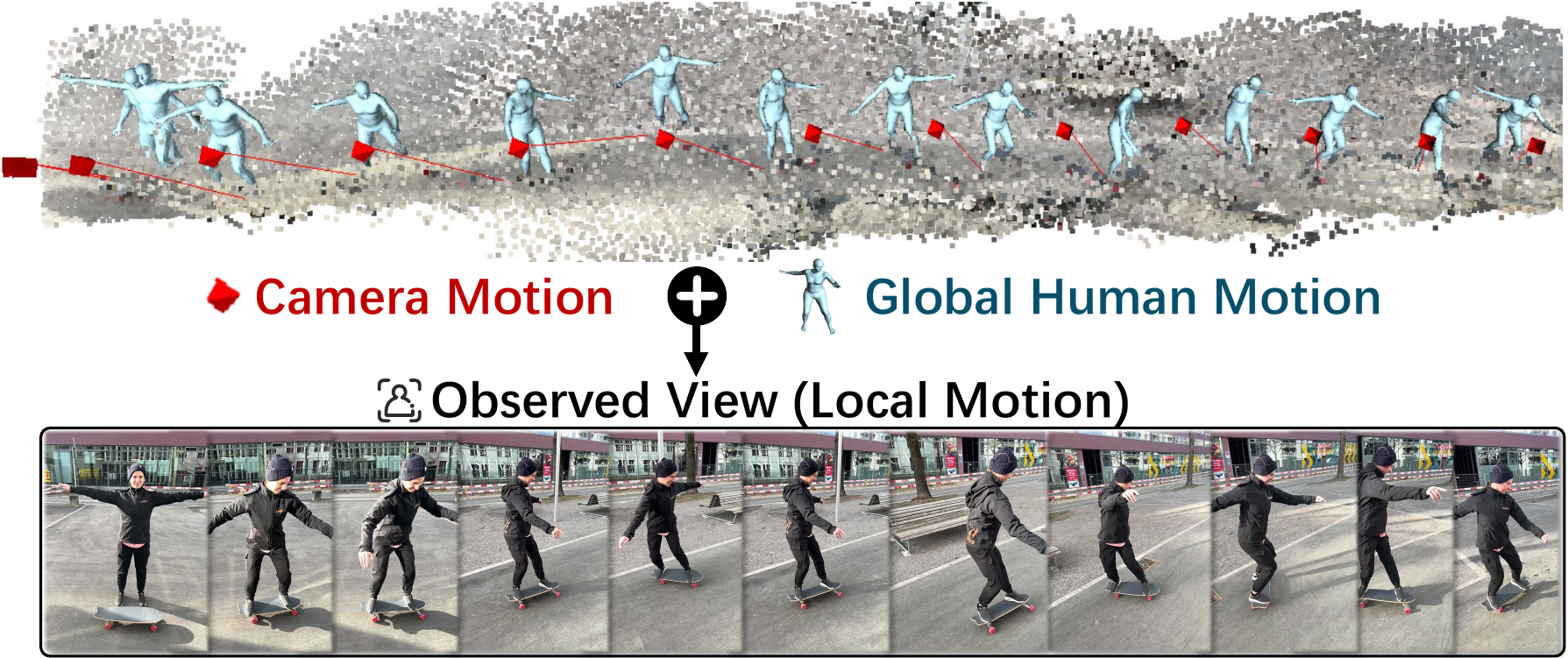}
        \caption{ \texorpdfstring{Camera $\circ$ Global = Local motion}{Camera+Human motion=Local motion}.} 
        \vspace{0.05cm}
        \label{fig:teaser:sketch}
    \end{subfigure}
    \\
    \begin{subfigure}[b]{0.236\textwidth}
        \centering
        \includegraphics[width=\textwidth]{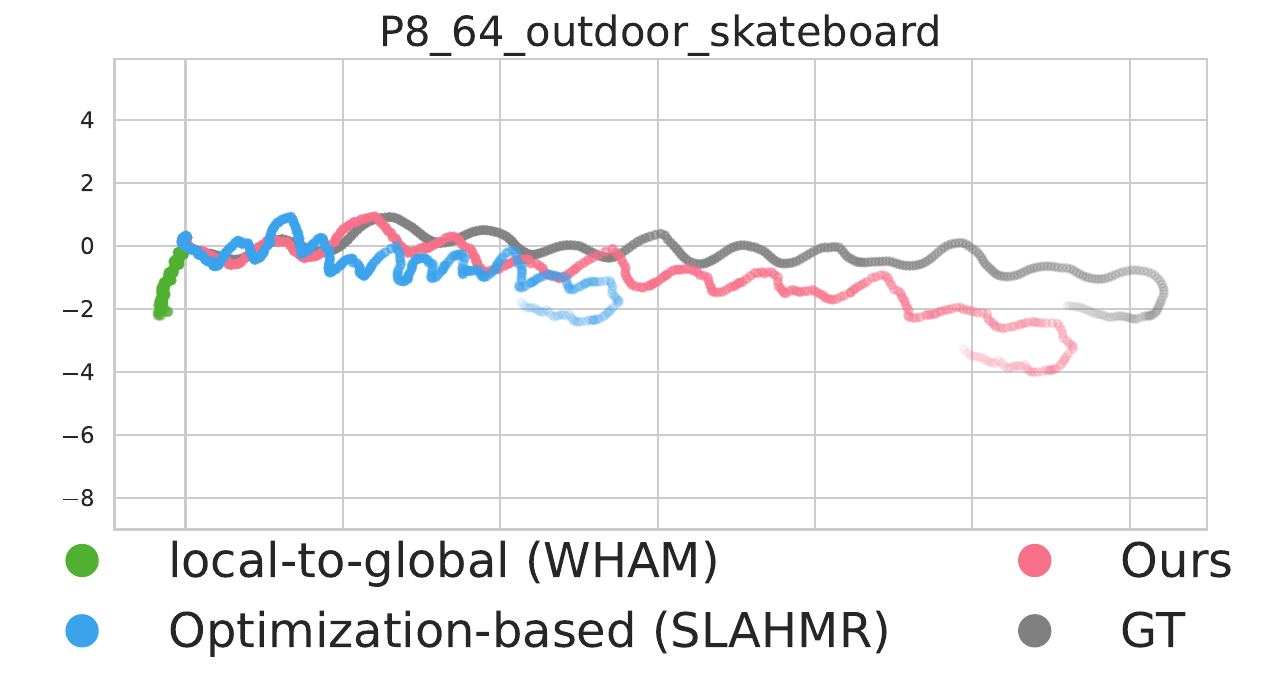}
        \caption{Human trajectories. } 
        \label{fig:teaser:traj}
    \end{subfigure}
    \begin{subfigure}[b]{0.236\textwidth}
        \centering
        \includegraphics[width=\textwidth]{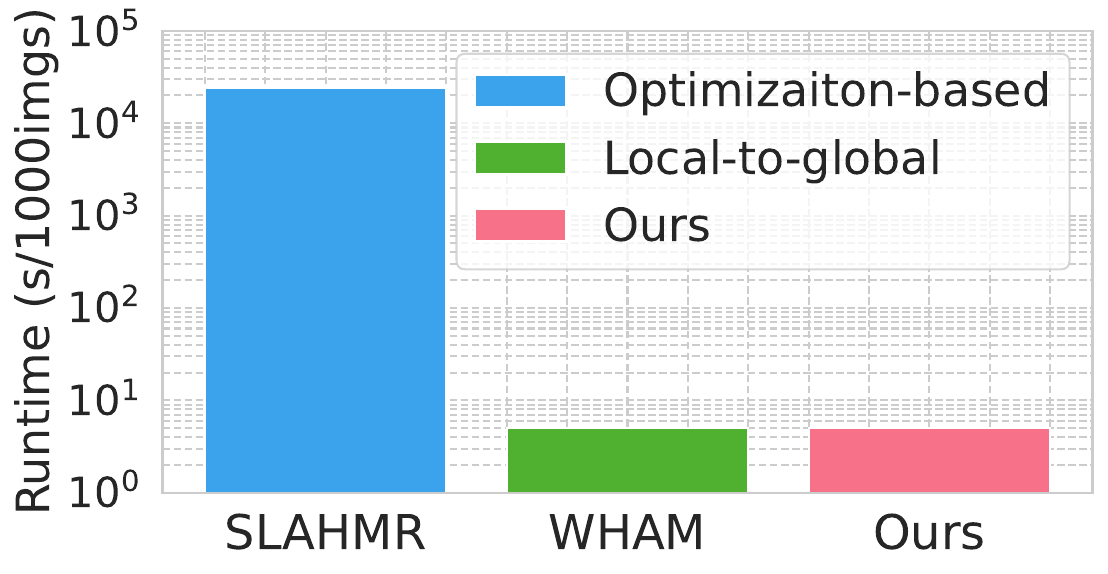}
        \caption{Inference time comparison. }
        \label{fig:teaser:opimization}
    \end{subfigure}
    \vspace{-0.7cm}
    \caption{(a) Video sequence as an entanglement of the camera and human motion in the world coordinate. (b) and (c) Local-to-global methods like WHAM~\cite{WHAM_2023} are time-efficient but fail in ambiguous cases; optimization-based methods like SLAHMR~\cite{SLAMHR_2023} struggle to optimize a good trajectory and are time-consuming. In contrast,  our method achieves accurate trajectories without optimization.}
    \label{fig:teaser}
    \vspace{-0.5cm}
\end{figure}

Monocular Human Mesh Recovery (HMR) is a key component of many vision applications~\cite{AR_rauschnabel2022what, VR_han2022virtual}. There are many successful monocular HMR methods~\cite{HMR_2017,VIBE_2020,SPIN_2019,TCMR_2021,CLIFF_2022,ImpHMR_2023,FAMI_2022}, but they are local, in that they can only estimate the mesh in the camera coordinates. Yet many vision applications may require the mesh in world, or global coordinates. Only a few world-coordinate or global HMR methods have been developed for monocular settings~\cite{GLAMR_2022, DnD_2022, TRACE_2023, WHAM_2023, TRAM_2024}.
Most learn a black-box local-to-global mapping directly from the local human motion\footnote{This work uses ``motion'' to refer to a sequence of human poses or meshes, or camera extrinsics over time.} pattern. 
Yet in some cases, motion ambiguity occurs when the background is ignored. For instance, a person gliding on a skateboard and someone standing still both exhibit similar local motion but differ significantly in global motion (see Fig.~\ref{fig:teaser:sketch}).

To resolve this ambiguity, it is essential to estimate camera motion from background information to isolate the true global human motion from local motion observed in video as shown in Fig.~\ref{fig:teaser:sketch}.
Monocular SLAM, commonly used for this purpose, estimates camera trajectories in RGB videos by tracking static key points and reconstructing a scene’s point cloud~\cite{SLAM_2006,DPVO_2023,DROIDSLAM_2021}. However, monocular SLAM only estimates camera motion and point cloud up to an unknown scale factor (see right part of Fig.~\ref{fig:framework}).  

Recent global HMR methods~\cite{SLAMHR_2023, PACE_2023} attempt to solve for the SLAM scale factor through optimization. The optimization is mainly based on a consistency loss between the 2D-projected human mesh and video evidence. The optimization jointly solves for the scale, human pose, and global motion.  
However, the entanglement of human and camera motion makes the optimization very challenging; some estimated scales have discrepancies up to 100\% or more (see SLAHMR~\cite{SLAMHR_2023} in Fig.~\ref{fig:teaser:traj}). Additionally, the optimizations are time-consuming; taking several minutes or longer to process a one-minute video (see Fig.~\ref{fig:teaser:opimization}).

In this paper, we propose to calibrate scale directly without the need for heavy optimization. In classic robotics, this is achieved by using known and fixed size patterns such as checkerboards~\cite{hartley2003multiple}. The underlying principle for scale calibration is inferring the metric depth of corresponding reference points within the point cloud from the known dimensions of objects like checkerboards. However, extra calibration tools are not available for human-centric videos.

Our key insight of this work is that humans in the scene can serve as calibration references to compute SLAM's scale factor. Despite their dynamic nature and size variability, we show that the absolute depth predicted by mature local HMR models is sufficiently precise to serve as the scale calibration reference. The metric depth prediction error of HMR models ($\sim0.1m$) is significantly smaller than the global human trajectory error ($>1m$), providing strong prior to estimate absolute scale.

To robustly link the metric depth of human joints to the relative depth of SLAM point cloud,  
we select reference points at or near human-background contacts (\eg, feet) and use the predicted contact joint depth from HMR models as the depth of this reference point to directly calibrate the SLAM scale. For scenarios where contact points between humans and the environment are non-visible (\eg, egocentric footage where feet and the floor are out of view), HMR can predict contact joint positions while the reference points might be missing from the scene point clouds. To address this, we estimate the ground plane where the reference points are expected, using the predicted human mesh to guide the orientation of the plane and ensure accuracy (\eg, contact joints are always on the ground plane).

In summary, we propose \textbf{H}uman \textbf{a}s \textbf{C}heckerboard (\textbf{HAC}), a novel framework that leverages humans as calibration references to accurately recover human mesh in world coordinates.  Our experimental results across diverse video conditions demonstrate significantly higher accuracy compared to existing methods. Furthermore, HAC reduces computational time by two to three orders of magnitude relative to optimization-based approaches (see Fig.~\ref{fig:teaser:opimization}).

We highlight our key contributions as follows:
\begin{itemize}
    \item The leveraging of humans in the scene as calibration references, based on strong metric depth cues from HMR models for direct and accurate scale calibration.
    \item An optimization-free scale calibration method, which significantly reduces computational time. 
    \item A ground plane fitting strategy to impute reference points for out-of-view contact joints. 
    \item Advanced results in global human and camera motion, also can effectively handle challenge ego-centric input.
\end{itemize}


\section{Related Works}
\label{sec:RelatedWorks}

\subsection{World Coordinate Human Mesh Recovery}
Image-based HMR methods~\cite{HMR_2017,BodyNet_2018,DenseBody_2019,I2l-meshnet_2020,pifuhd_2020,lintransformers_2021,Meshgraphormer_2021,rong2019delving,sengupta2020synthetic,ImpHMR_2023} mostly 
recover human meshes within the camera’s coordinate system. Many video-based HMR advancements~\cite{VIBE_2020, TCMR_2021, FAMI_2022} are also based in 
the camera space, though introducing 
video data has paved the way for HMR exploration in world coordinates~\cite{GLAMR_2022, DnD_2022, bodyslam_2022, SLAMHR_2023, TRACE_2023, WHAM_2023}. Transitioning to world coordinates introduces the challenge of disentangling dynamic camera motion and human motion. Some attempts (\eg, GLAMR~\cite{GLAMR_2022}, DnD~\cite{DnD_2022}, TRACE~\cite{TRACE_2023}, and WHAM~\cite{WHAM_2023}) infer global motion from observable local behaviors, \eg, a person who looks to be walking is assumed to be moving forwards globally.  However, this local-to-global mapping is inherently ambiguous, \eg, the person may be walking on a treadmill.
In contrast, we explicitly model camera motion and decouple global human motion from local observations to eliminate such ambiguities. 

%


Other works SLAHMR~\cite{SLAMHR_2023} and PACE~\cite{PACE_2023} utilize background information in determining camera motion with SLAM techniques~\cite{ORBSLAM_2015, DROIDSLAM_2021, DPVO_2023}. These methods address the `unknown scale' problem by jointly optimizing scale, pose, and shape parameters, which can be unstable and computationally intensive. 
In contrast, our optimization-free method directly calibrates the scale factor using depth predictions from HMR models.

Recently, TRAM~\cite{TRAM_2024} recovers scale by using depth predictions from off-the-shelf metric depth estimators~\cite{Zoedepth_2023} onto the background scene. In contrast, our HAC method uses foreground human depth from HMR models for scale calibration. This approach offers improved robustness in challenging scenarios, such as large camera movements and limited background information.

\subsection{Camera Scale Calibration}
Camera scale calibration is a fundamental procedure in 3D computer vision for ensuring real-world accuracy of the camera motion.
Traditional methods use reference markers like chekcerboards~\cite{hartley2003multiple} or additional sensors like IMUs~\cite{zhang2000flexible, heikkila1997four} to determine a known metric scale.
However, these methods are not suitable for human-centric videos without sensor measurements or additional calibration tools. 
Single-view metrology methods~\cite{singleviewmetrology_2020, criminisi2000single, hoiem2008putting, kar2015amodal} have shown that objects with well-defined geometric priors can act as natural calibration references. 
Inspired by this, we use human predictions from HMR models as surrogate calibration references, with a proof-of-concept presented in a subsequent section. Our approach uses the absolute joint depths from HMR to accurately and efficiently calibrate the unknown scale factor.

\section{Preliminaries}
\label{sec:Preliminaries}

\subsection{Human Motion in Camera Coordinates \texorpdfstring{$\mathcal{M}_c$}{Mc}}
The 3D human motion from a video ${\{I_t\}}_{t=1}^T$ of $T$ frames can be represented in the camera space by a $T$-length sequence of SMPL parameters $\mathcal{M}_c = {\{\boldsymbol{\theta}_t, \boldsymbol{\beta}_t, \boldsymbol{\psi}_t, \boldsymbol{\tau}_t\}}_{t=1}^{T}$, where
SMPL~\cite{SMPL_2015} is a widely used 3D statistical model of the human body.  
For a given frame at time $t$, the SMPL model maps body pose $\boldsymbol{\theta}_t \in \mathbb{R}^{23 \times 3}$, shape parameters $\boldsymbol{\beta}_t \in \mathbb{R}^{10}$, root orientation $\boldsymbol{\psi}_t \in \mathbb{R}^{3}$, and root translation $\boldsymbol{\tau}_t \in \mathbb{R}^{3}$ to a 3D mesh of the human body $ \mathbf{V}\!\in\!\mathbb{R}^{6890 \times 3} $ in the camera space.  The individual joints can be mapped from the SMPL parameters with the function $\mathbf{J}\!=\!\mathcal{J}(\boldsymbol{\theta}_t, \boldsymbol{\beta}_t, \boldsymbol{\psi}_t, \boldsymbol{\tau}_t)\!\in\!\mathbb{R}^{24 \times 3}$. 
Note that $\boldsymbol{\psi}_t$ and $\boldsymbol{\tau}_t$ are sometimes referred to as ``global'' orientation and translation parameters by the SMPL model.  However, HMR models~\cite{HMR_2017, SPIN_2019, VIBE_2020, TCMR_2021, PARE_2021, vitpose_2022} estimate these parameters with respect to the camera extrinsics 
$\mathcal{E}$ at frame $t$, where $\mathcal{E} = {\{{\boldsymbol{R}}_t, {\boldsymbol{T}}_t\}}_{t=1}^{T}$ is the sequence of camera rotations $ {\boldsymbol{R}}_t \in \mathbb{R}^{3 \times 3} $ and translations ${\boldsymbol{T}}_t \in \mathbb{R}^{3}$. 

\subsection{Human Motion in World Coordinates \texorpdfstring{$\mathcal{M}_w$}{Mw}}
Unlike $\mathcal{M}_c$, the human motion in the world coordinates $\mathcal{M}_w = {\{\boldsymbol{\theta}_t, \boldsymbol{\beta}_t, \boldsymbol{\Phi}_t, \boldsymbol{\Gamma}_t\}}_{t=1}^{T}$ is the motion within an absolute global space \footnote{By convention, the world space is defined by the camera extrinsics parameters of the very first frame.} and independent of camera extrinsics $\mathcal{E}_t$. Recall the classical perspective projection, local root orientation and translation $\{\boldsymbol{\psi}_t, \boldsymbol{\tau}_t\}$ is obtained by applying camera extrinsics $\mathcal{E}_t$ to the global orientation $\boldsymbol{\Phi}_t \in \mathbb{R}^{3}$ and global translation $\boldsymbol{\Gamma}_t \in \mathbb{R}^{3}$ in world coordinates:
\begin{equation} \label{eg:world-to-camera}
\boldsymbol{\psi}_t=\boldsymbol{R}_t \boldsymbol{\Phi}_t; \quad \boldsymbol{\tau}_t=\boldsymbol{R}_t \boldsymbol{\Gamma}_t + \boldsymbol{T}_t.
\end{equation}
To obtain the global human motion from the local motion estimated by HMR models, one can apply the inverse of the camera extrinsics to the local root: 
\begin{equation} \label{eg:camera-to-world}
\boldsymbol{\Phi}_t=\boldsymbol{R}_t^\top \boldsymbol{\psi}_t ; \quad \boldsymbol{\Gamma}_t=\boldsymbol{R}_t^\top (\boldsymbol{\tau}_t - \boldsymbol{T}_t).
\end{equation}
This equation decouples the global motion from the local estimation by isolating and removing the camera motion.  It serves as one of our key insights. 

\subsection{SLAM} \label{sec:slam}
Nevertheless, accurately estimating camera extrinsics 
can be difficult. Simultaneous Localization and Mapping (SLAM) ~\cite{ORBSLAM_2015, DROIDSLAM_2021, DPVO_2023} is widely used to predict the camera extrinsics 
$\mathcal{E}' = {\{\boldsymbol{R}_t,\boldsymbol{T'}_t\}}_{t=1}^{T}$ and scene point clouds $\mathcal{P} = \{ \mathbf{x'}_i \in \mathbb{R}^3 \}_{i=1}^N $, where $T'$ and $x'$ are camera translations and point cloud in SLAM system, up to an unknown scale factor $s$; $N$ is the number of points in $\mathcal{P}$.
Our work focuses on calibrating the scale $s$ by using recovered human mesh as the `checkerboard'. Similar to those traditional calibration strategies, this work estimates $s$ based on the ratio of absolute versus relative distance to the camera for some reference point $p$ in the reconstructed scene, \ie 
\begin{equation} \label{eq:scale}
    s_p = d^{A}_{p} / d^{S}_{p},
\end{equation}
where $d^{A}_{p}$ and $d^{S}_{p}$ denote the absolute real-world distance from camera to $p$ and the corresponding relative distance in SLAM, respectively. We will show details of how to determine these two distances in the later section.


\subsection{Ground Plane Fitting} \label{sec:fitground}
Given the scene point cloud $\mathcal{P}$, our goal is to identify the plane $\{\mathbf{n}, b\}$ that best represents the ground surface, where $\mathbf{n}, b$ are the normal vector and offsets respectively. While standard methods like RANSAC-based plane segmentation can effectively fit the largest plane, it struggles with complex scenes where the largest plane is not necessarily the ground (\eg walls or objects). In such cases, prior knowledge about the typical orientation and position of the ground plane is needed (\eg horizontal and at the lowest height cluster). In this work, we leverage human mesh as guides to determine the ground plane.


\section{Method}
\label{sec:method}
\begin{figure*}[t!]
    \vspace{-0.3cm}
	\centering
	\includegraphics[width=1.\linewidth]{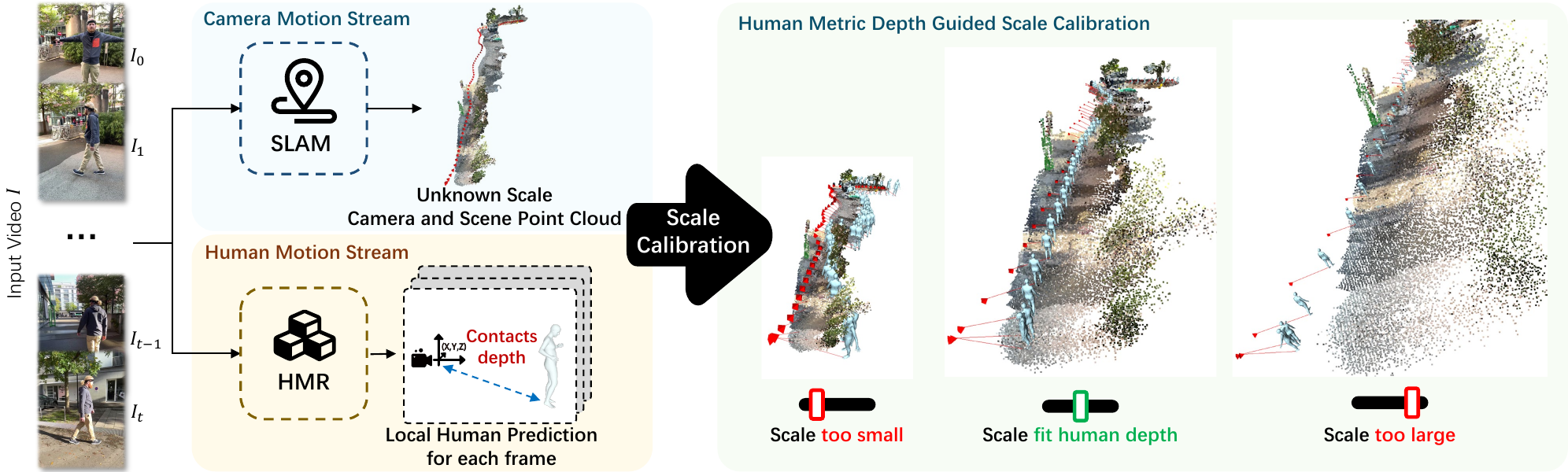}
    \vspace{-0.5cm}
    \caption{Overall pipeline of HAC. Given a monocular input video, we use SLAM to estimate camera motion and scene reconstruction at an arbitrary scale. Concurrently, we predict local human mesh using an HMR model. We then use the metric depth of contact joints from HMR to calibrate the SLAM scale. With this approach, we can accurately decouple global human and camera motion in world coordinates.
    }

	\label{fig:framework}
    \vspace{-0.3cm}
\end{figure*}

In this section, we first provide a proof-of-concept for using humans as checkerboards (\cref{subsec:proof-of-concept}). We then describe how human-scene contact joints can be used to calibrate SLAM scale (\cref{subsec:ScaleEstimator}). For frames without visible contact joints, we introduce a ground plane estimation strategy (\cref{subsec:non-visiblecontact}). Finally, we determine a robust scale factor for the entire video, so as to get the final camera motion and global human motion (\cref{subsec:scaleforvideo}). We illustrate our proposed pipeline in~\cref{fig:framework}.

\subsection{Validity of Using Humans as Checkerboards} \label{subsec:proof-of-concept}

\begin{figure}[h!]
    \vspace{-0.3cm}
	\centering
	\includegraphics[width=1.0\linewidth]{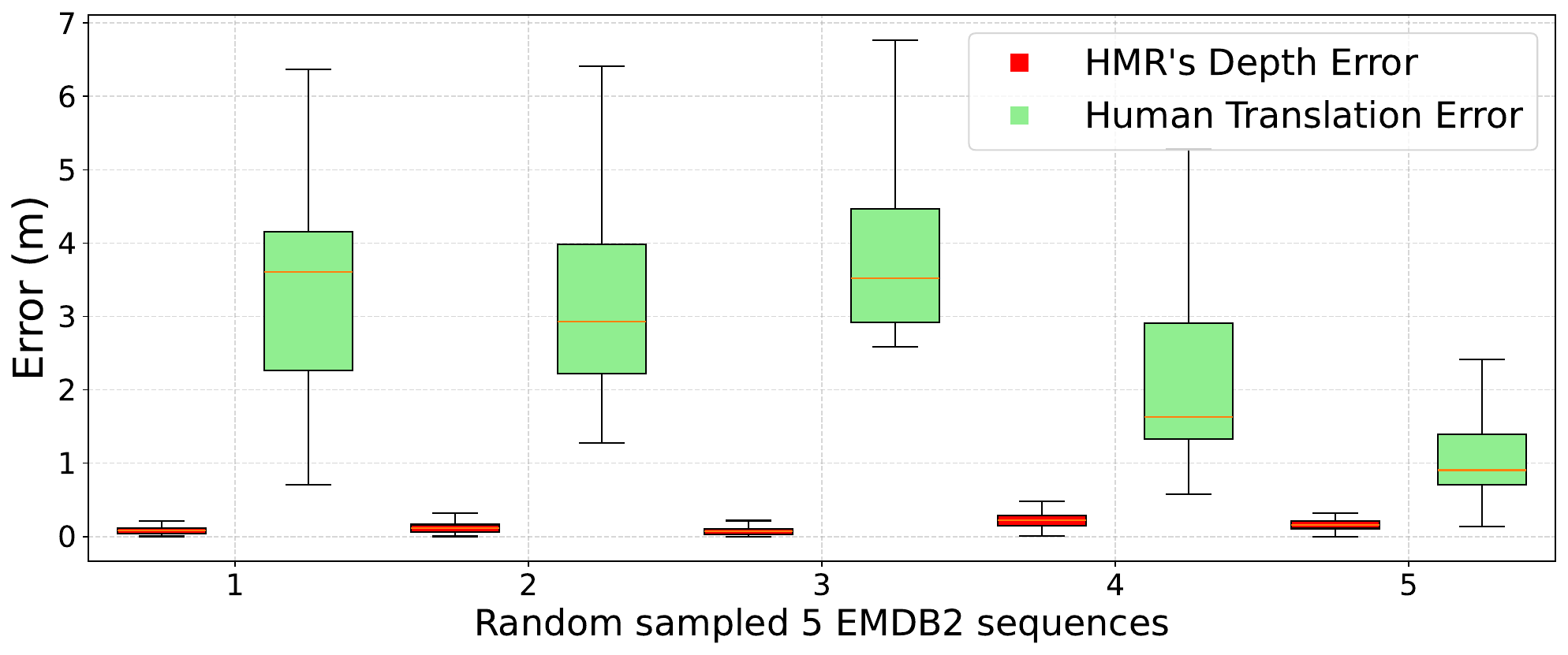}
    \vspace{-0.8cm}
	\caption{Error rate boxplot on the HMR predicted metric depth and the global human translation estimation by WHAM~\cite{WHAM_2023}. We show that HMR depth consistently exhibits a significantly lower error rate compared to global translation, which validates our concept of using humans as checkerboards for scale calibration.
 }
	\label{fig:proofofconcept}
    \vspace{-0.2cm}
\end{figure}

As existing HMR models are trained on large-scale datasets~\cite{PW3D_2018, human36m_2013, AMASS_2019,COCO_2014, MPI_INF_3DHP_2017} and use statistical models like SMPL~\cite{SMPL_2015}, they can provide strong geometric cues such as predicting the metric depth of human joints. 
As shown in ~\cref{fig:proofofconcept}, the predicted depth error from HMR ($\sim0.1m$) is significantly smaller than the global human trajectory error ($>1m$) encountered in our task, validating HMR predictions as reliable metric information for scale calibration. We provide additional details in Sec. {A} of the Supplementary.


\begin{figure*}[tbp]
    \vspace{-0.3cm}
	\centering
	\includegraphics[width=1.\linewidth]{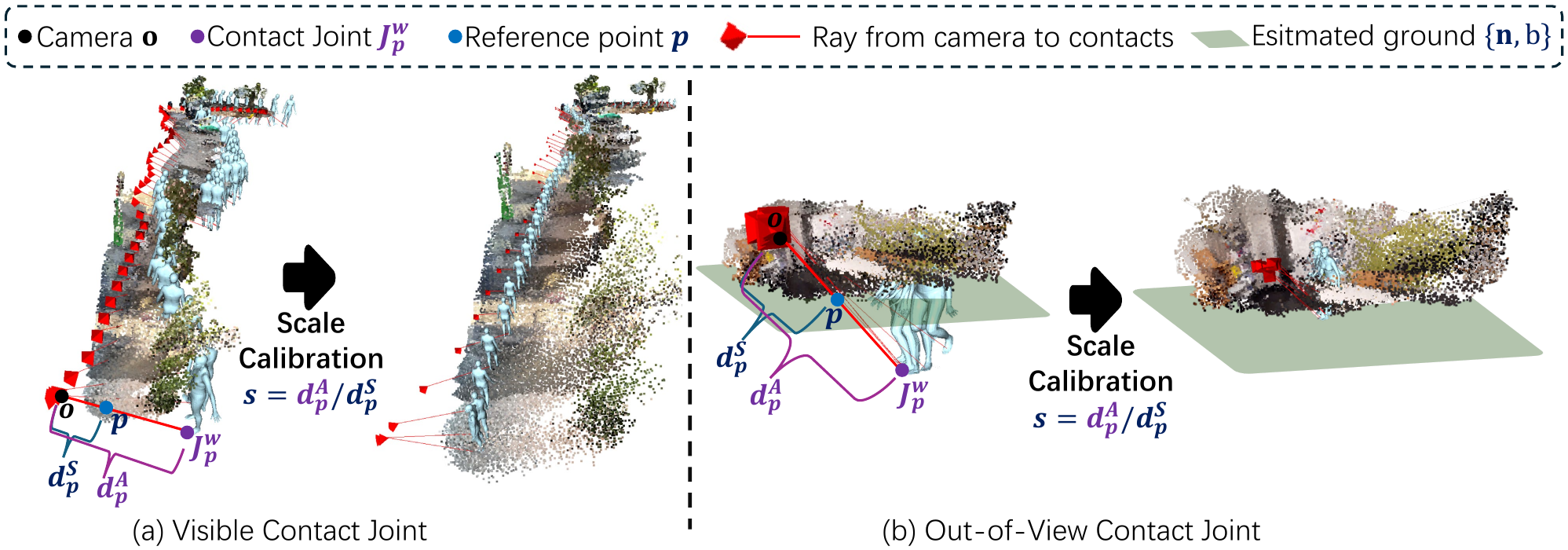}
    \vspace{-0.3cm}
	\caption{Detail of our scale calibration process. The scale is calibrated by comparing the depth from the camera $\mathbf{o}$ to the contact joint $\mathbf{J}^{w}_{p}$ predicted by HMR with the depth to the corresponding reference point $p$ in the scene point cloud. There are two possible scenarios for the reference point $p$: (a) when the contact joint is visible, $p$ is obtained as the intersection with the point cloud; (b) when the contact joint is not visible (\eg, no intersection or incorrect intersection), $p$ is determined as the intersection with the estimated ground plane. 
 }
	\label{fig:scalecalib}
    \vspace{-0.5cm}
\end{figure*}

\subsection{Scale Estimation Using Human-Scene Contacts} \label{sec:scalevisible}
\label{subsec:ScaleEstimator}


As discussed in Sec.~\ref{sec:slam}, our goal is to calibrate unknown scale $s$ by the ratio of absolute distance $d^{A}_{p}$ versus relative distance $d^{S}_{p}$ to the camera  for some reference point $p$ in $\mathcal{P}$. In this section, we show how we select the reference point $p$ and calculate corresponding $d^{A}_{p}$ and $d^{S}_{p}$ for calibrating $s$ for each frame by Eq.~\ref{eq:scale}.

{\noindent\textbf{Human-scene Contacts as Reference Points.}
To leverage the strong metric cues provided by HMR models, we need to robustly link the metric depth of human joints to the relative depth of the corresponding points in the SLAM point cloud.
However, SLAM typically masks out dynamic objects including humans, and only tracks key points in the static background. Therefore, as illustrated in Fig.~\ref{fig:scalecalib}, we propose to use the human-scene contacts as the anchor for calibration, where $\mathbf{J}_p$ is the contact joint in camera coordinate predicted by HMR, and $p$ is its corresponding reference point located at SLAM point cloud. As such, the HMR predicted depth of joint $\mathbf{J}_p$ can be a good estimate of the absolution depth $d^{A}_{p}$, while the distance between $p$ and the SLAM predicted camera $o$ is the relative depth $d^{S}_{p}$. 


}

\noindent\textbf{Absolute Distance $d^{A}_{p}$ Derived from HMR Model.}
We assume that the lowest human joint along the y-axis of the camera coordinate system\footnote{In this work, a y-down coordinate system is used, where the y-axis points downward and the z-axis aligns with the camera’s viewing direction.} is the human-scene contact joint 
$ \mathbf{J}_{p} = \arg\max_{\mathbf{J}_{i} \in \mathbf{J}}  \mathbf{J}_{i}^{(y)}$. 
By definition, the camera is the origin of the camera space, so the absolute distance from contact joint $\mathbf{J}_{p}$ to the camera is given by $d^{A}_{p} =  \left\| \mathbf{J}_{p} \right\|_2.$


\noindent\textbf{Relative Distance $d^{S}_{p}$ Derived from SLAM.}
We can then determine the reference point $p \in \mathcal{P}$ corresponding to the contact joint $\mathbf{J}_p$. 
Recall in Sec.~\ref{sec:slam}, SLAM methods estimate the scene point cloud $\mathcal{P}$ and camera motion ${\{\boldsymbol{R}_t, \boldsymbol{T'}_t\}}_{t=1}^{T}$. 
As shown in Fig.~\ref{fig:scalecalib}(a), the reference point $p$ is the intersection of the point cloud $\mathcal{P}$ and the ray $\mathbf{r}$ originating from the camera $\mathbf{o}=\boldsymbol{R}_t^\top (-\boldsymbol{T'}_t)$, directed toward the contact joint $\mathbf{J}_p^w$ in SLAM's world coordinate:
\begin{equation} 
p = \arg \min_{\mathbf{p}_i \in \mathcal{P}} \left\| (\mathbf{p}_i - \mathbf{o}) - ((\mathbf{p}_i - \mathbf{o}) \cdot \hat{\mathbf{d}}) \hat{\mathbf{d}} \right\|_2.
\end{equation}
Here, $\hat{\mathbf{d}} = \frac{\mathbf{J}_p^w - \mathbf{o}}{\|\mathbf{J}_p^w - \mathbf{o}\|}$ is the normalized direction vector of the ray $\mathbf{r}$. $\mathbf{J}_p^w$ is the contact joint in SLAM's world coordinate, which is obtained by applying the camera-to-world transformation to $\mathbf{J}_p$ in camera space, as illustrated in Eq.~\ref{eg:camera-to-world}. After finding reference point $p$, we can obtain the relative distance $d^{S}_{p}$: 
\begin{equation} \label{eq:dslam} 
    d^{S}_{p} = \left\| p - \mathbf{o} \right\|_2.
\end{equation}

By substituting $d^{A}_{p}$ and $d^{S}_{p}$ into Eq.~\ref{eq:scale}, we can obtain the scale $s$ for this specific frame $t$.

\subsection{Out-of-View Contact Joints.}
\label{subsec:non-visiblecontact}
In most cases, the reference point $p$ can be found as the intersection of the scene point cloud $\mathcal{P}$ and the ray $\mathbf{r}$. However, there are situations where no valid intersection exists (\eg, due to incomplete floor reconstruction in certain egocentric views), or the intersection is incorrect (\eg, due to obstructions from furniture or other objects in the scene.). These cases generally occur when the contact joint is not visible. To handle such cases, we fit an approximate ground plane to estimate the relative depth as in Fig.~\ref{fig:scalecalib}(b). Recall in Sec.~\ref{sec:fitground}, we utilize the HMR prediction as the guide determining the norm vector of the ground, and then the offset of the ground is fitted using the SLAM point cloud, ensuring consistency between HMR predictions and scene geometry.

\noindent\textbf{Determining the Ground Plane Normal from HMR.}
We fit the ground plane by assuming all contact joints across the video are on the ground floor. As shown in previous works~\cite{SLAMHR_2023,Humor_2021}, the ground plane normal vector $\mathbf{n}$ can be calculated by minimizing a plane's distance to all contact joints across frames in the world coordinate $\{{\mathbf{J}_p^w}_t \in \mathbb{R}^3\}_{t=1}^T$. This can be solved by Least Squares Fitting.
An alternative way is to simply set the normal vector $\mathbf{n}$ as the y-axis of the SLAM's system, which is less robust as shown in our ablation studies in Tab.~\ref{tab:nonvisible}.

\noindent\textbf{Determining Ground Plane Offset from SLAM.}
With the normal vector $\mathbf{n}$, we segment the SLAM point cloud $\mathcal{P}$ along the normal vector and use the lowest segment $\mathcal{P}_l \subset \mathcal{P}$ to determine the ground plane offset $b$. The optimal $b$ is found such that the ground plane $\{\mathbf{n}, b\}$ passes through as many points as possible in the lowest segment. 
Alternatively, one could fit the largest possible plane within the point cloud as mentioned in Sec.~\ref{sec:fitground}. Still, this approach may not always be feasible, due to incomplete floor reconstruction in certain scenarios, as also demonstrated in our ablation study in Tab.~\ref{tab:nonvisible}.

\noindent\textbf{Relative Distance Derived from SLAM.}
After obtaining the ground plane $\{\mathbf{n}, b\}$, the reference point $p$ is the intersection of the ray $\mathbf{r}$ with this plane:
\begin{equation}
p = \mathbf{o} + t^* \hat{\mathbf{d}}, \quad \text{where } t^* = -\frac{b + \mathbf{n}^\top \mathbf{o}}{\mathbf{n}^\top \hat{\mathbf{d}}}.
\end{equation}
Thus, the relative distance $d^{S}_{p}$ can still be calculated as described in Eq.~\ref{eq:dslam}. Same as in Sec.~\ref{sec:scalevisible}, the scale for the current frame $t$ can be calculated by Eq.~\ref{eq:scale}.

\subsection{Final Human Motion and Camera Motion}
\label{subsec:scaleforvideo}

Taking into account the scales calculated across all $T$ frames, we can determine the overall scale for the entire sequence using various measures like the mean or median. It is important to note that the scale for some single frames can be noisy due to inconsistencies in SLAM estimates and HMR predictions. Therefore, to mitigate the impact of noise and ensure robustness, we use the median of the scale factors across all frames of the video as the final scale factor: $\bar{s} = \text{median}(\{s(t) \mid t \in I\})$ for the whole sequence $I$, ensuring stability and robustness.

After calibrating the scale as $\bar{s}$, the absolute camera extrinsics is ${\{\boldsymbol{R}_t, \bar{s} \cdot \boldsymbol{T'}_t\}}_{t=1}^{T}$. By applying them to HMR's predicted human motion $\mathcal{M}_c$, we can finally get the global human  motion:
\begin{equation}
    \mathcal{M}_w = {\{\boldsymbol{\theta}_t, \ \boldsymbol{\beta}_t, \ \boldsymbol{\Phi}_t=\boldsymbol{R}_t^\top \boldsymbol{\psi}_t, \ \boldsymbol{\Gamma}_t=\boldsymbol{R}_t^\top (\boldsymbol{\tau}_t - \bar{s} \cdot \boldsymbol{T'}_t)\}}_{t=1}^{T},
\end{equation}
where $\boldsymbol{\theta}_t, \boldsymbol{\beta}_t$ are from HMR's prediction, and $\boldsymbol{\Phi}_t, \boldsymbol{\Gamma}_t$ are calculated by camera-to-world translation (\ie Eq.~\ref{eg:camera-to-world}) from HMR's prediciton $\boldsymbol{\psi}_t, \boldsymbol{\tau}_t$ in camera space.


\begin{table*}[t!]
    \vspace{-0.4cm}
    \centering
    \caption{Ablation studies of our methods on EMDB dataset. There are three ablation sections on rows: (1) the impact when no camera motion at all, only using local motion to predict global motion~\cite{GLAMR_2022}; (2) the impact if without scale calibration, directly using arbitrary scale from the original SLAM output; (3) the impact of using different reference joints for scale calibration.}
    \vspace{-0.3cm}
\begin{tabular}{l|ccc|c}
\hline
\multirow{2}{*}{Ablation}                                    & \multicolumn{3}{c|}{Global Human Motion}                     & Global Camera Motion \\ \cline{2-5} 
                                                             & W-MPJPE$\downarrow$ & WA-MPJPE$\downarrow$ & RTE$\downarrow$ & ATE-S$\downarrow$    \\ \hline
w/o camera motion~\cite{GLAMR_2022} & 788.50  & 293.05 & 7.72  & -    \\ \hline
w/o scale calibration               & 949.75  & 396.09 & 10.74 & 7.25 \\ \hline
Head fix as $\mathbf{J}_{p}$        & 1069.97 & 372.86 & 8.96  & 8.68 \\
Pelvis fix as $\mathbf{J}_{p}$      & 772.33  & 289.76 & 8.02  & 6.37 \\
Feet fix as $\mathbf{J}_{p}$        & 198.94  & 71.94  & 1.30  & 0.82 \\ \hline
\rowcolor{lightgray!30} \textbf{Contact as $\mathbf{J}_{p}$} & \textbf{197.24}     & \textbf{70.95}       & \textbf{1.22}   & \textbf{0.79}        \\ \hline
\end{tabular}
    \label{tab:ablation}
    \vspace{-0.3cm}
\end{table*}

\begin{table}[t!]
\vspace{-0.1cm}
\caption{Ablation study on ground plane fitting strategies in scenarios with non-visible contact joint. Numbers in grey indicate the upper bound achieved using the ground truth ground plane.}
\vspace{-0.3cm}
\resizebox{1.0\linewidth}{!}{
\begin{tabular}{l|ccc}
\hline
\multirow{2}{*}{Fit ground plane}   & \multicolumn{3}{c}{EgoBody}                                  \\ \cline{2-4} 
                                    & W-MPJPE$\downarrow$ & WA-MPJPE$\downarrow$ & RTE$\downarrow$ \\ \hline
w/o fitting                         & 146.6               & 87.8                 & 2.0             \\
standard fitting                 & 144.3               & 87.9                 & 2.0             \\
y-axis fix as $\mathbf{n}$          & 118.9               & 78.0                 & 1.2             \\
\textcolor{gray}{upper-bound}                          & \textcolor{gray}{106.3}               & \textcolor{gray}{73.0}                 & \textcolor{gray}{0.9}             \\ \hline
\rowcolor{lightgray!30}  \textbf{$\{\mathbf{n}, b\}$ (ours)} & \textbf{109.8}      & \textbf{74.8}        & \textbf{0.9}    \\ \hline
\end{tabular}
}
\label{tab:nonvisible}
\vspace{-0.5cm}
\end{table}

\section{Experiments}
\label{sec:experiments}
\subsection{Implementation Details, Dataset and Metrics}
Our method\footnote{The code will be released upon acceptance} supports mainstream HMR models, including CLIFF~\cite{CLIFF_2022}, HMR2.0~\cite{HMR2_2023}, and VIMO~\cite{TRAM_2024} for the human motion stream; and mainstream SLAM methods such as DroidSLAM~\cite{DROIDSLAM_2021}, DPVO~\cite{DPVO_2023} for the camera motion stream. We use the most advanced configurations of VIMO and the masked DroidSLAM for the full system. 
We use a RANSAC strategy to fit the ground plane normal vector. 

\noindent \textbf{Datasets.} Following~\cite{WHAM_2023}, 
we evaluate the global human motion and camera motion on a subset of EMDB~\cite{EMDB_2023} (EMDB 2), which contains 25 sequences captured by a dynamic camera and provides ground truth global motion for both the human and the camera.  We also evaluate on EgoBody~\cite{EgoBody_2022}, an ego-centric dataset, with sequences captured by a head-mounted camera. The videos feature significant body truncations, providing a challenging testbed for evaluating our methodology for out-of-view contact joints.

\noindent \textbf{Metrics for Human Motion.} 
We follow previous works~\cite{PACE_2023, SLAMHR_2023, WHAM_2023} and evaluate human's global motion error by: (1) \textbf{WA-MPJPE} (mm), the average Euclidean distance between the ground truth and the predicted joint positions (\ie MPJPE) after aligning each segment for every 100 frames; (2) \textbf{W-MPJPE} (mm), the MPJPE error after only aligning the first two frames of each 100-frames segments with the ground truth; and (3) \textbf{RTE} (\%), the human's root translation error normalized by the actual displacement of the whole sequence after the rigid alignment. We also evaluate local mesh error by (4) \textbf{PA-MPJPE}  (mm), the MPJPE error after Procrustes alignment with the ground truth. 

\noindent \textbf{Metrics for Camera Motion.}
We follow the SLAM convention~\cite{ATE_2012, PACE_2023} and evaluate by: (1) \textbf{ATE} (m), the Average Translation Error after rigidly aligning the camera trajectories; and (2) \textbf{ATE-S} (m), the Average Translation Error without scale alignment, providing a more accurate reflection of inaccuracies in the captured scale of the scene.

\begin{table*}[t!]
    \vspace{-0.3cm}
    \centering
    \caption{SOTA comparison for global and local human motion on the EMDB and EgoBody dataset. $^*$ denotes the errors we achieve from TRAM's HMR model release. 
    }
    \vspace{-0.3cm}
    \resizebox{1.0\linewidth}{!}{
    \begin{tabular}{l|cccc|cccc}
\hline
\multirow{2}{*}{Models}   & \multicolumn{4}{c|}{EMDB2}   & \multicolumn{4}{c}{EgoBody}    \\ \cline{2-9} 
 &
  W-MPJPE$\downarrow$ &
  WA-MPJPE$\downarrow$ &
  PA-MPJPE$\downarrow$ &
  RTE$\downarrow$ &
  W-MPJPE$\downarrow$ &
  WA-MPJPE$\downarrow$ &
  PA-MPJPE$\downarrow$ &
  Runtime/1000imgs \\ \hline
GLAMR~\cite{GLAMR_2022}   & 726.6  & 280.8 & 56.0 & 16.7 & 416.1 & 239.0 & 114.3 & 7min   \\
TRACE~\cite{TRACE_2023}   & 1702.3 & 529.0 & 58.0 & 18.9 & -     & -     & -     & -      \\
SLAHMR~\cite{SLAMHR_2023} & 776.1  & 326.9 & 61.5 & 10.2 & 141.1 & 101.2 & 79.13 & 400min \\
PACE~\cite{PACE_2023}     & -      & -     & -    & -    & 147.9 & 101.0 & 66.5  & 8min   \\ 
WHAM~\cite{WHAM_2023}     & 354.8  & 135.6 & 41.9 & 6.0  & -     & -     & -     & -      \\ 
TRAM~\cite{TRAM_2024}     & 222.4  & 76.4  & 38.1 & 1.4  & -     & -     & -     & -      \\ 
\rowcolor{lightgray!30} \textbf{HAC (Ours)} &
  \textbf{197.2} &
  \textbf{71.0} &
  {38.6$^*$} &
  \textbf{1.2} &
  \textbf{109.8} &
  \textbf{74.8} &
  \textbf{46.8} &
  \textbf{5sec} \\ \hline
\end{tabular}
    }
    \label{tab:SOTA}
    \vspace{-0.1cm}
\end{table*}

\begin{table*}[tbp]
\caption{Comparison of camera motion error and corresponding global human motion error across sequence lengths in EMDB2: short ($<$40 sec), medium ($<$60 sec), and long ($>$60 sec). The first row presents results using an off-the-shelf metric depth predictor~\cite{Zoedepth_2023} as the depth input. The subsequent rows detail the performance of scale recovery methods, TRAM and HAC (ours), for monocular SLAM output.
}
\vspace{-0.3cm}
\resizebox{1.0\linewidth}{!}{
\begin{tabular}{l|ccc|ccc|ccc}
\hline
\multirow{2}{*}{Methods} & \multicolumn{3}{c|}{Short}          & \multicolumn{3}{c|}{Medium}        & \multicolumn{3}{c}{Long}           \\ \cline{2-10} 
                        & ATE$\downarrow$  & ATE-S$\downarrow$ & W-MPJPE$\downarrow$ & ATE$\downarrow$  & ATE-S$\downarrow$ & W-MPJPE$\downarrow$ & ATE$\downarrow$  & ATE-S$\downarrow$ & W-MPJPE$\downarrow$ \\ \hline
masked DroidSLAM + ZoeDepth~\cite{Zoedepth_2023} & 0.68 & 0.94   &  309.57  & 2.13 &  2.39  & 305.89    & 5.10 & 5.53  & 408.83   \\
mono masked DroidSLAM + TRAM's scale & \textbf{0.24} & 0.51          & 283.54          & \textbf{0.25} & 0.62          & 231.83          & \textbf{0.47} & \textbf{0.80} & 218.97          \\
\rowcolor{lightgray!30} \textbf{mono masked DroidSLAM + Ours scale} & \textbf{0.24} & \textbf{0.28} & \textbf{258.51} & \textbf{0.25} & \textbf{0.48} & \textbf{180.13} & \textbf{0.47} & 1.09          & \textbf{195.30} \\ \hline
\end{tabular}
}
\label{tab:scalevideolength}
\vspace{-0.1cm}
\end{table*}

\begin{figure*}[t!]
    \vspace{-0.1cm}
	\centering
	\includegraphics[width=1.\linewidth]{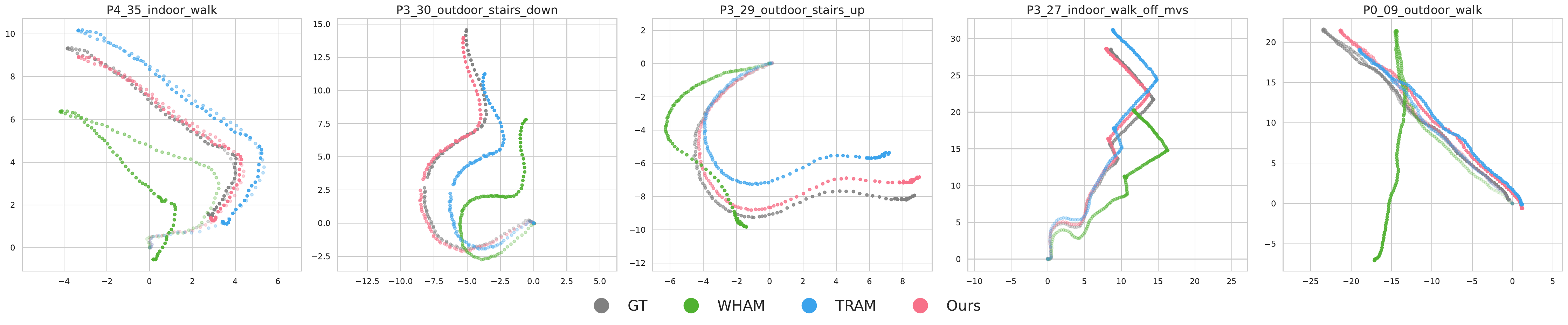}
    \vspace{-0.8cm}
	\caption{Comparison of global human trajectory estimation on EMDB. Overall, ours shows better alignment to ground truth data compared to WHAM and TRAM.}
	\label{fig:humantraj}
    \vspace{-0.5cm}
\end{figure*}

\subsection{Ablation Study and Analysis} \label{sec:ablation}
\textbf{Importance of Camera Motion.}
Directly predicting the global motion from local motion priors~\cite{GLAMR_2022, DnD_2022, TRACE_2023} can be ambiguous.  For example, GLAMR~\cite{GLAMR_2022} (`w/o camera motion' in Tab.~\ref{tab:ablation}) has high errors compared to modeling the camera motion as we do (`Contact as $\textbf{J}_p$' in the last row).


\noindent \textbf{Scale Calibration.}
The initial scale of SLAM outputs is far from accurate; fixing the scale to the initial scale estimated by the SLAM output from masked DroidSLAM is even worse (Tab.~\ref{tab:ablation}, `w/o scale calibration') than the estimation by local-to-global HMR method~\cite{GLAMR_2022}.
Our scale calibration (see Tab.~\ref{tab:ablation}, `Contact as $\textbf{J}_p$') effectively estimates global motions for both the human (left panel) and the camera (right panel).  
In Sec. B of the Supplementary, we provide examples of the camera trajectory before and after our scale calibration. The initial SLAM output underestimates the scale, whereas our calibrated estimates align much more closely with the ground truth.


\noindent \textbf{Reference Joint Selection.}
The third section of Tab.~\ref{tab:ablation} highlights the higher errors when 
selecting non-contact joints such as the head or pelvis as reference points. 
Using the feet as the reference (fifth row) 
is almost as good as selecting the lowest joint in each frame as the contact joint (last row) since the feet are 
the contact joints in the majority of the frames. We provide additional analysis on reference joints in Sec. C of the Supplementary.

\noindent \textbf{Non-visible Contact Joint Scenarios.}
In challenging scenarios like the EgoBody dataset, the contact joint is not visible for 94\% of the time (no contacts seen at all for some sequences). Without fitting the ground plane (first row in Tab.~\ref{tab:nonvisible}), the performance degrades noticeably compared to our proposed approach (last row). Standard fitting by segmenting the largest plane (second row) is also ineffective, as the largest detected plane is often a wall or another non-floor structure given a noisy point cloud. Alternatively, aligning the SLAM system's y-axis to the normal vector (third row) provides moderate improvement by utilizing the rough alignment of the SLAM system's y-axis with gravity direction, offering a useful prior for plane orientation. However, our approach (last row), which leverages human contact joints as a strong prior for plane orientation, achieves near upper-bound performance where ground-truth plane is using (fourth row). The remaining error is primarily due to noise in human pose predictions caused by severe truncation in the EgoBody dataset.


\subsection{Comparison with the State-of-the-art}
\begin{figure*}[t!]
    \vspace{-0.3cm}
	\centering
	\includegraphics[width=1.0\linewidth]{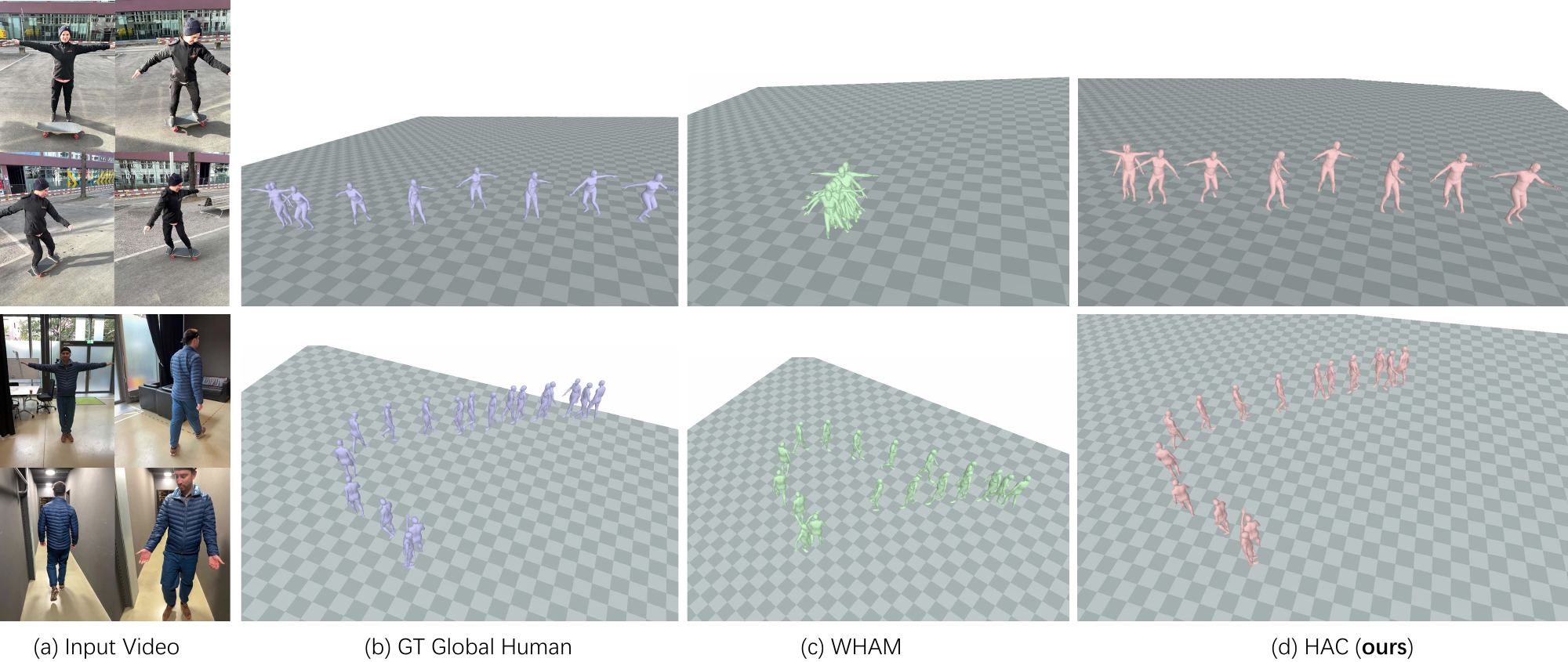}
    \vspace{-0.7cm}
	\caption{
    Qualitative examples of the proposed approach. We show that HAC can recover global human motions more accurately than the previous state-of-the-art method (c) WHAM~\cite{WHAM_2023}. We provide additional video examples and scene reconstruction in the Supplementary.
    }
	\label{fig:vislization}
    \vspace{-0.3cm}
\end{figure*}

As shown in Tab.~\ref{tab:SOTA}, HAC significantly improves global human motion metrics over the previous SOTA~\cite{TRAM_2024} by 15\% improvement in W-MPJPE, 10\% improvement in WA-MPJPE on EMDB2 dataset. We present additional qualitative examples in Fig.~\ref{fig:vislization}, demonstrating that our method produces more plausible global human motion when compared with the ground truth.

\noindent \textbf{Human Global Translation Ambiguity.} The large global trajectory error of local-to-global methods (see RTE of TRACE, GLAMR, and also WHAM in Tab.~\ref{tab:SOTA}) demonstrates the difficulties they face when attempting to infer global trajectories solely from local information. As shown in the first row in Fig.~\ref{fig:vislization}, it's hard to infer the global translation from a "standing" local pose when a woman is skateboarding (green meshes). By explicitly modeling camera motion to decouple human motion from the local observation, our method achieves a 50\% improvement over the previous local-to-global SOTA method WHAM, resulting in a significantly more accurate human trajectories (see Fig.~\ref{fig:humantraj}).

\noindent \textbf{Time Complexity.} The last column in Tab.~\ref{tab:SOTA} compares the runtime between our method and scale-optimization methods (such as SLAHMR~\cite{SLAMHR_2023} and PACE~\cite{PACE_2023}) for 1000 frames. All methods incorporate SLAM, so we report runtimes excluding SLAM to fairly compare only the optimization components. SLAHMR takes around 400 minutes for optimization, while PACE is much faster and needs 8 minutes. Our approach is about 100x even faster, requiring only 5 seconds on a single RTX 2080 Ti GPU. Beyond better time efficiency, HAC also delivers better performance in scale estimation and global human trajectory accuracy. as shown in Fig.~\ref{fig:teaser:traj} and Tab.~\ref{tab:SOTA}.

\noindent \textbf{Comparison with TRAM on Scale Calibration.}
Tab.~\ref{tab:scalevideolength} compares the accuracy of camera motion after scale calibration using our method (HAC) and TRAM, showing that HAC consistently outperforms TRAM across almost all scenarios. 
The primary reason is that TRAM relies on off-the-shelf metric depth prediction~\cite{Zoedepth_2023} which is unreliable in scenes with limited amount of background information or large camera motion. 
Although our human-aware scale calibration may be affected by HMR's depth noise, our scale factor effectively adjusts to this bias, ensuring that the calibrated camera motion remains consistent with the human movement. For instance, a slightly larger scale factor can offset a higher predicted depth, preserving alignment between the calibrated camera and human motion (more details in Sec. A of the Supplementary). This ability to adapt to prediction noise is why our method achieves better global human performance even in cases where ATE-S might be slightly higher for longer sequences. 

\subsection{Compatibility Analysis} \label{sec:moreanalysis}
Our method leverages human priors from HMR to calibrate SLAM scale and is compatible with various off-the-shelf HMR and SLAM approaches.

\noindent \textbf{Different Local HMR Models.}
Tab.~\ref{tab:virousHMR} shows that our method supports all mainstream HMR models and consistently outperforms TRAM across these models. When comparing CLIFF (first segment) and HMR2.0 (second segment), it shows better local pose does not guarantee better global motion. This limitation arises because precise local pose estimates do not inherently ensure depth accuracy or temporal consistency in poses. Furthermore, video-based HMR VIMO (last segment) outperforms image-based models like CLIFF and HMR2.0 by leveraging temporal dynamics to achieve superior local and global results.

\begin{table}[t!]
\centering
\vspace{-0.1cm}
\caption{Comparison of scale calibration performance across different HMR models, showing ours consistent better than TRAM. PA-, WA-, and W- denote corresponding MPJPE metrics.}
\vspace{-0.3cm}
\resizebox{0.9\linewidth}{!}{
\begin{tabular}{l|l|cccc}
\hline
\multirow{2}{*}{HMR} & \multirow{2}{*}{Scale} & \multicolumn{4}{c}{EMDB2} \\ \cline{3-6} 
 &  & PA-$\downarrow$ & WA-$\downarrow$ & W-$\downarrow$ & RTE$\downarrow$ \\ \hline
\multirow{2}{*}{CLIFF~\cite{CLIFF_2022}} & TRAM & \multirow{2}{*}{53.7} & 89.8 & 255.6 & 1.3 \\
 & \cellcolor{lightgray!30} \textbf{Ours} &  & \cellcolor{lightgray!30} \textbf{83.4} & \cellcolor{lightgray!30}\textbf{232.1} & \cellcolor{lightgray!30}\textbf{1.2} \\ \hline
\multirow{2}{*}{HMR2.0~\cite{HMR2_2023}} & TRAM & \multirow{2}{*}{48.5} & 89.0 & 279.3 & 1.4 \\
 & \cellcolor{lightgray!30} \textbf{Ours} &  & \cellcolor{lightgray!30} \textbf{85.1} & \cellcolor{lightgray!30}\textbf{260.7} & \cellcolor{lightgray!30}\textbf{1.3} \\ \hline
\multirow{2}{*}{VIMO~\cite{TRAM_2024}} & TRAM & \multirow{2}{*}{38.6} & 79.7 & 230.9 & 1.4 \\
 & \cellcolor{lightgray!30} \textbf{Ours} &  & \cellcolor{lightgray!30} \textbf{71.0} & \cellcolor{lightgray!30}\textbf{197.2} & \cellcolor{lightgray!30}\textbf{1.2} \\ \hline
\end{tabular}
}
\label{tab:virousHMR}
\vspace{-0.1cm}
\end{table}

\begin{table}[t!]
\vspace{-0.1cm}
\caption{Effectiveness of our calibration across different SLAM methods. Numbers in grey indicate the upper bound achieved using the ground truth camera motion.}
\vspace{-0.3cm}
\resizebox{1.0\linewidth}{!}{
\begin{tabular}{ll|ccc}
\hline
\multicolumn{1}{l|}{SLAM} & Scale & WA$\downarrow$ & W-MPJPE$\downarrow$ & RTE$\downarrow$ \\ \hline
\multicolumn{1}{l|}{\multirow{2}{*}{DroidSLAM~\cite{DROIDSLAM_2021}}} & w/o calib. & 422.8 & 1327.3 & 14.9 \\
\multicolumn{1}{l|}{} & \cellcolor{lightgray!30}\textbf{Ours} & \cellcolor{lightgray!30}\textbf{187.6} & \cellcolor{lightgray!30}\textbf{614.9} & \cellcolor{lightgray!30}\textbf{6.8} \\ \hline
\multicolumn{1}{l|}{\multirow{2}{*}{DPVO~\cite{DPVO_2023}}} & w/o calib. & 331.9 & 826.6 & 9.6 \\
\multicolumn{1}{l|}{} & \cellcolor{lightgray!30}\textbf{Ours} & \cellcolor{lightgray!30}\textbf{95.8} & \cellcolor{lightgray!30}\textbf{316.6} & \cellcolor{lightgray!30}\textbf{2.8} \\ \hline
\multicolumn{1}{l|}{\multirow{2}{*}{masked DroidSLAM~\cite{TRAM_2024}}} & w/o calib. & 396.1 & 949.7 & 10.7 \\
\multicolumn{1}{l|}{} & \cellcolor{lightgray!30}\textbf{Ours} & \cellcolor{lightgray!30}\textbf{71.0} & \cellcolor{lightgray!30}\textbf{197.2} & \cellcolor{lightgray!30}\textbf{1.2} \\ \hline
\multicolumn{2}{l|}{\textcolor{gray}{GT Camera}} & \textcolor{gray}{65.7} & \textcolor{gray}{189.7} & \textcolor{gray}{0.2} \\ \hline
\end{tabular}
}
\label{tab:virousSLAM}
\vspace{-0.3cm}
\end{table}

\noindent \textbf{Different SLAM Methods.}
Tab.~\ref{tab:virousSLAM} demonstrates the compatibility and effectiveness of HAC's calibration across various SLAM methods.  Masked DroidSLAM (third section), which masks out dynamic humans in the foreground, is more stable than DroidSLAM (first section) and DPVO (second section), which use the original video and are affected by foreground dynamics. Notably, our current performance closely approaches the upper bound using ground truth camera motion (last row), suggesting the remaining errors are primarily due to limitations in human pose accuracy and consistency over time.

\section{Limitations}

Like other methods that rely on SLAM~\cite{SLAMHR_2023, PACE_2023, TRAM_2024}, one limitation of our approach is that HAC inherently dependent on the performance of the SLAM system. In catastrophic scenarios where SLAM fails, our method is unable to produce accurate results. Additionally, SLAM drift can occur when processing very long video sequences, introducing noise into the estimated global human motion. 

\section{Conclusion}

This paper proposes Humans as Checkerboards (HAC), which uses HMR's absolute depth prediction as a tool to calibrate the unknown scale of SLAM. By utilizing human-scene contacts as the calibration reference, HAC effectively and efficiently recovers the camera motion. This allows our method to accurately decouple global human motion from video observations without expensive optimization used by previous methods—simply placing single-frame pose estimates in 3D space is sufficient to achieve state-of-the-art results. Our experiments demonstrate HAC's robustness across diverse and challenging video conditions.


{
    \small
    \bibliographystyle{ieeenat_fullname}
    \bibliography{main}
}

\clearpage
\setcounter{page}{1}
\maketitlesupplementary

\setcounter{section}{0}
\renewcommand{\thesection}{\Alph{section}}

\section{Validity of Humans as Checkerboards} \label{sec:supp:validity}
In this section, we provide additional details to support the concept of using humans as scale calibration references, as discussed in Sec.~\textcolor[rgb]{.5,.5,1}{4.1} of the main paper.

\begin{figure}[h!]
    \vspace{-0.3cm}
    \centering
    \begin{subfigure}[b]{0.49\textwidth}
        \centering
        \includegraphics[width=\textwidth]{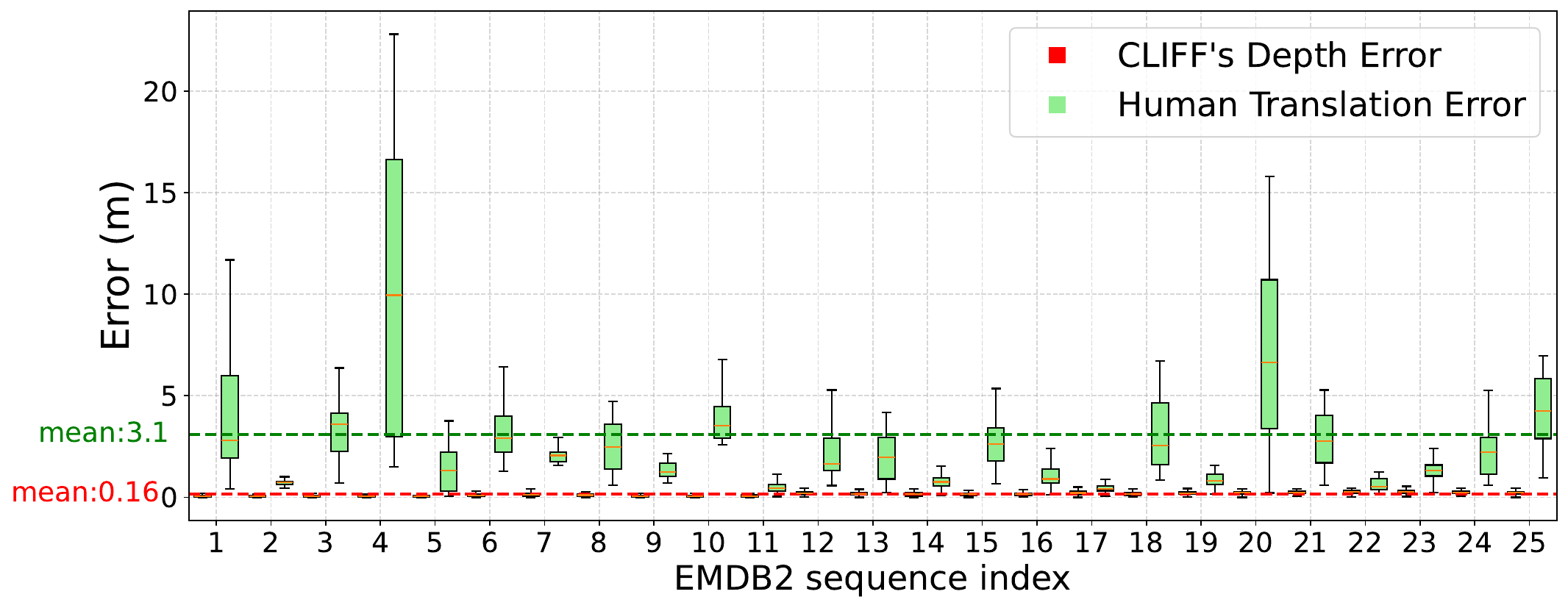}
        \vspace{-0.5cm}
        \caption{CLIFF's depth prediction error versus global human translation error.} 
        \label{fig:validity:samllerr:cliff}
    \end{subfigure}
    \vspace{0.02cm}
    \\
    \begin{subfigure}[b]{0.49\textwidth}
        \centering
        \includegraphics[width=\textwidth]{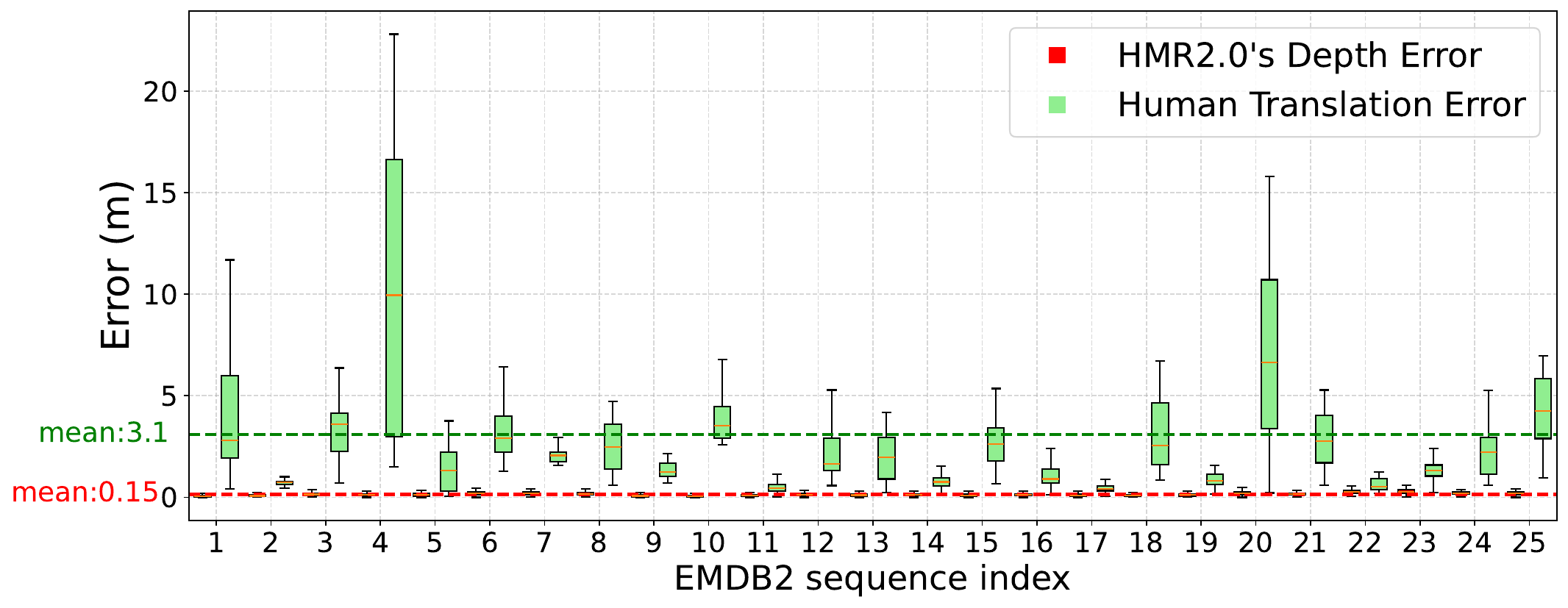}
        \vspace{-0.5cm}
        \caption{HMR2.0's depth prediction error versus global human translation error.} 
        \label{fig:validity:samllerr:hmr2}
    \end{subfigure}
    \vspace{-0.5cm}
    \caption{Quantitative analysis of depth prediction errors from different HMR models, (a) CLIFF and (b) HMR2.0, compared to the global human trajectory error across the entire EMDB2 dataset.}
    \label{fig:validity:samllerr}
    \vspace{-0.2cm}
\end{figure}

\noindent \textbf{Sufficient Small Depth Error.}
Fig.~\textcolor[rgb]{.5,.5,1}{2} in the main paper showcased data from only five randomly sampled examples for better clarity and visualization, here in Fig.~\ref{fig:validity:samllerr} we expand this analysis by comparing the two errors across the entire EMDB2~\cite{EMDB_2023} dataset. 
Fig.~\ref{fig:validity:samllerr:cliff} demonstrates that the depth prediction error of CLIFF model~\cite{CLIFF_2022} is consistently lower than that of human translation across all samples. Additionally, as illustrated in Fig.~\ref{fig:validity:samllerr:hmr2}, testing with different HMR models such as HMR2.0~\cite{HMR2_2023} has shown a similar trend. These results show that HMR can predict sufficiently accurate depth compared to global translation error we are addressing in our task~\cite{WHAM_2023}. This validates our motivation for using humans as reliable metric cues for scale calibration.

\begin{figure}[h!]
    \vspace{-0.3cm}
	\centering
	\includegraphics[width=1.0\linewidth]{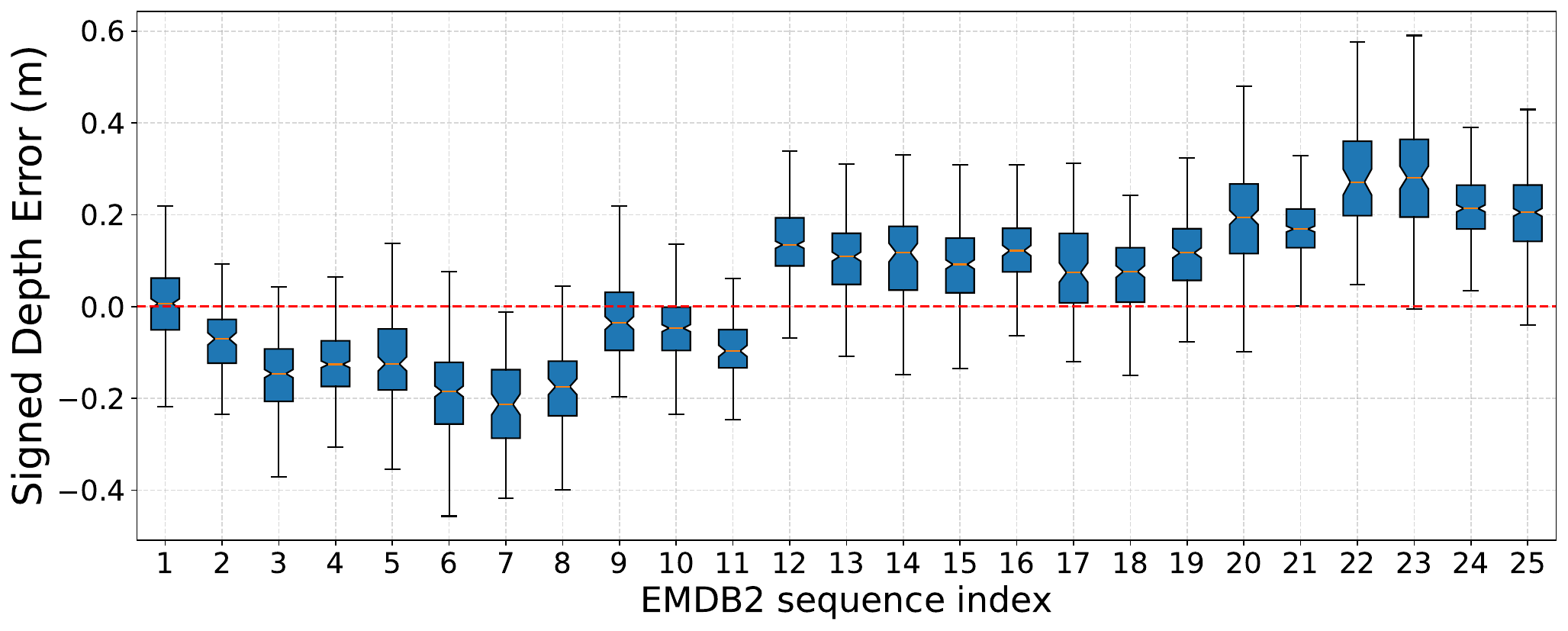}
    \vspace{-0.8cm}
	\caption{HMR tends to predict either predominantly larger or smaller depth for each video compared with ground truth.}
	\label{fig:validity:errorbias}
    \vspace{-0.55cm}
\end{figure}

\noindent\textbf{Consistent Bias in Depth Error.}
Besides the sufficient small error magnitude, the depth prediction errors from HMR show a consistent pattern within each video sequence. Compared with ground truth, HMR tends to predict either predominantly larger or smaller depth across all frames in a video (as shown in Fig.~\ref{fig:validity:errorbias}). This consistency is advantageous for calibration purposes: using humans as calibration references can systematically compensate for this bias. Our human-aware calibration effectively adapts to the inherent HMR depth bias, enabling precise overall global human motion recovery, as shown in Sec.~\textcolor[rgb]{.5,.5,1}{5.3} of the main paper.

\section{Qualitative Results for Scale Calibration}
\begin{figure*}[ht!]
	\centering
	\includegraphics[width=1.\linewidth]{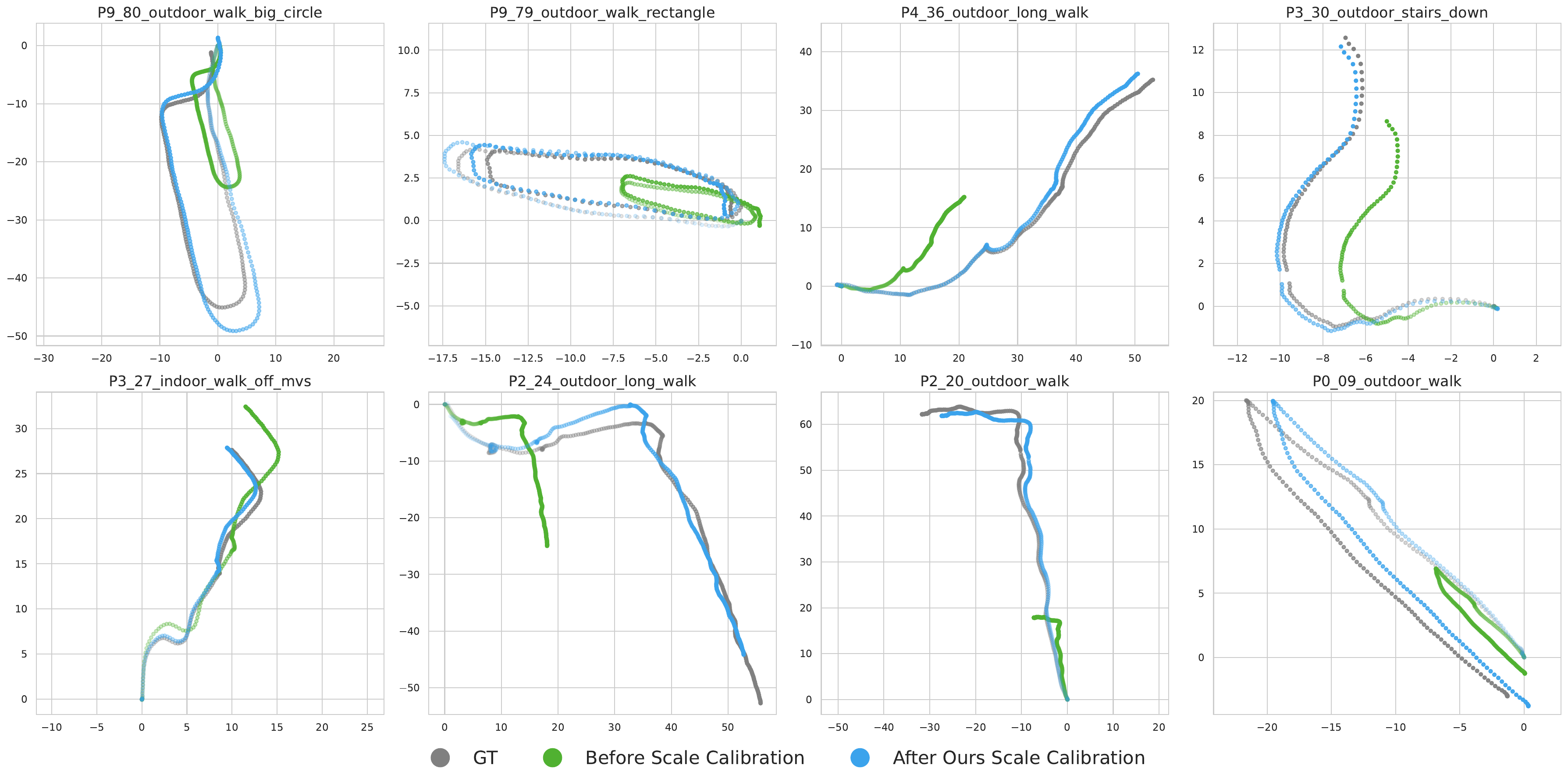}
    \vspace{-0.6cm}
	\caption{Visulizaiton of camera trajectory before and after our scale calibration. As shown, the original SLAM output is up to an unknown scale. After our scale calibration, it becomes better aligned to ground truth data.}
	\label{fig:camtraj}
    \vspace{0.2cm}
\end{figure*}

In this section, we present additional qualitative results that demonstrate the effectiveness of our scale calibration, as mentioned in Sec.~\textcolor[rgb]{.5,.5,1}{5.2} of the main paper. Fig.~\ref{fig:camtraj} illustrates the challenge of the unknown scale factor in the initial SLAM output (green trajectory). After our scale calibration, the result camera trajectory (blue trajectory) aligns more closely with the ground truth (grey trajectory), demonstrating the effectiveness of HAC.

\section{More Insights on Reference Joint Selection} 
\begin{figure*}[tbp]
    \vspace{0.3cm}
	\centering
	\includegraphics[width=1.0\linewidth]{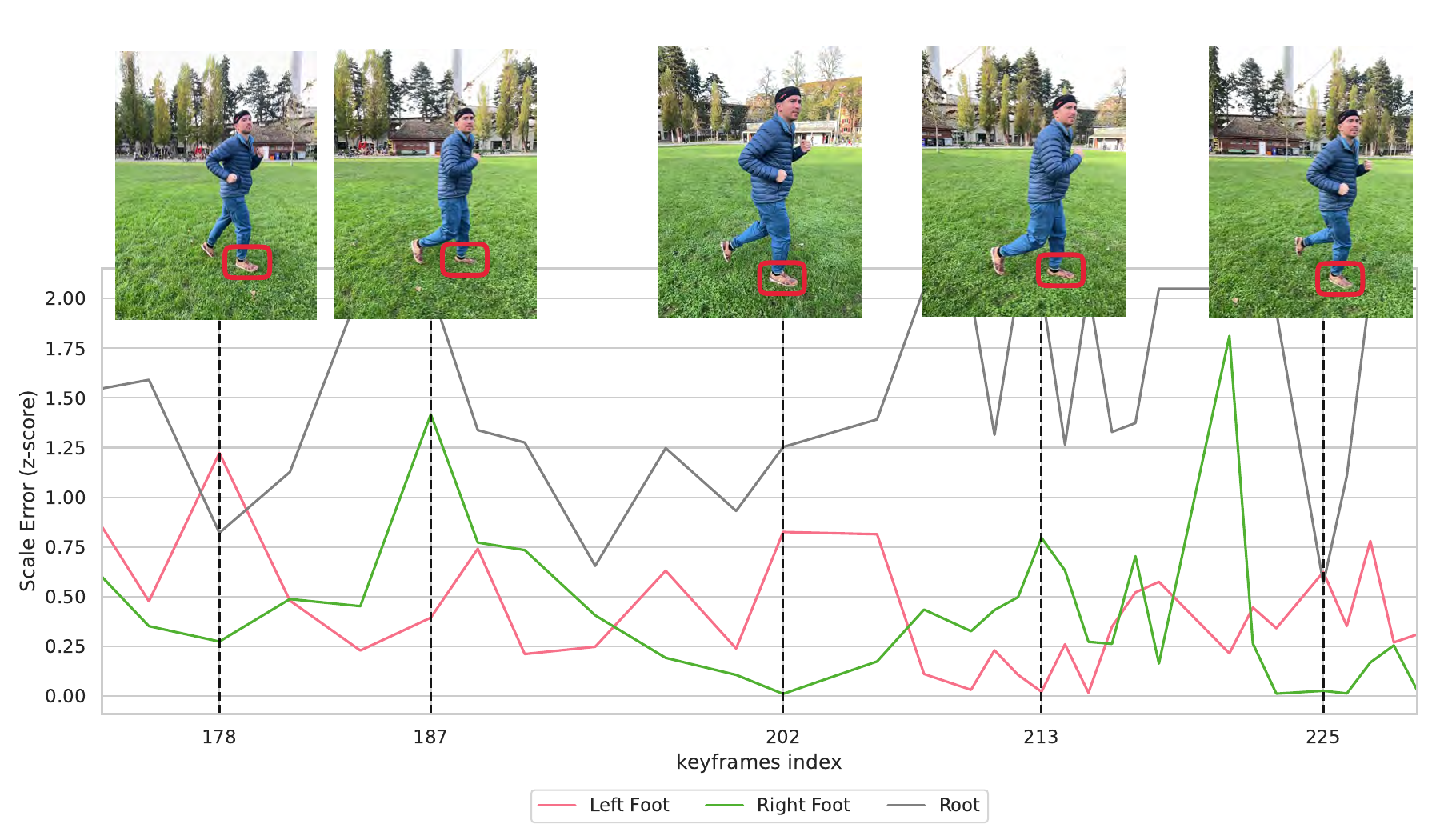}
    \vspace{-0.8cm}
	\caption{Scale error across some keyframes of the video. Results show a low scale error when the foot is in contact with the ground. Additionally, an inverse correlation between the left and right foot scale errors corresponds with the alternating pattern of footfalls during locomotion.}
	\label{fig:contactjoints}
\end{figure*}

In this section, we delve deeper into how the selection of reference joints impacts the performance of scale calibration, as discussed in Sec.~\textcolor[rgb]{.5,.5,1}{5.2} of the main paper.
Fig~\ref{fig:contactjoints} demonstrates pelvis joints show larger scale error than foot joints, aligning with the findings in Tab.~\textcolor[rgb]{.5,.5,1}{1} of the main paper. Furthermore, an inverse correlation exists between scale errors of left and right feet, with growth in left feet corresponding to a contraction in right. This scale error trend is also consistent with the contact feet as shown in the corresponding image frames, a foot will have less scale error when it contacts with the ground. This further verified our motivation for choosing the human-scene contact joint as the reference point.

\section{Video Demos and Point Clouds} 
In this supplementary folder, we provide the video \textbf{\textit{HAC\_Qualitative\_Examples.mp4}}, which illustrates some of our results in video form. Additionally, a folder named \textbf{\textit{scene\_reconstruction\_with\_human}} contains corresponding files that show the human and camera meshes along with the scene point clouds.

\textbf{\textit{HAC\_Qualitative\_Examples.mp4}} presents a three-panel side-by-side illustration: (i) the input video, (ii) our results from the current SLAM camera point of view, and (iii) our results from a global static point of view. These visualizations demonstrate that our reconstructed scale is accurate, with smooth and coherent human and camera motion fitting very well within the world coordinates.

\textbf{\textit{scene\_reconstruction\_with\_human} folder} contains the comprehensive human, camera meshes, and scene point cloud corresponding to the aforementioned video. For optimal viewing, we recommend using a 3D visualization tool such as MeshLab. Open \textit{*-scene\_point\_cloud.ply} and \textit{*-human\_and\_camera\_meshes.ply} together in MeshLab to explore the human and camera motion within the scene.


\section{Discussion}
\begin{figure}[h!]
	\centering
	\includegraphics[width=1.0\linewidth]{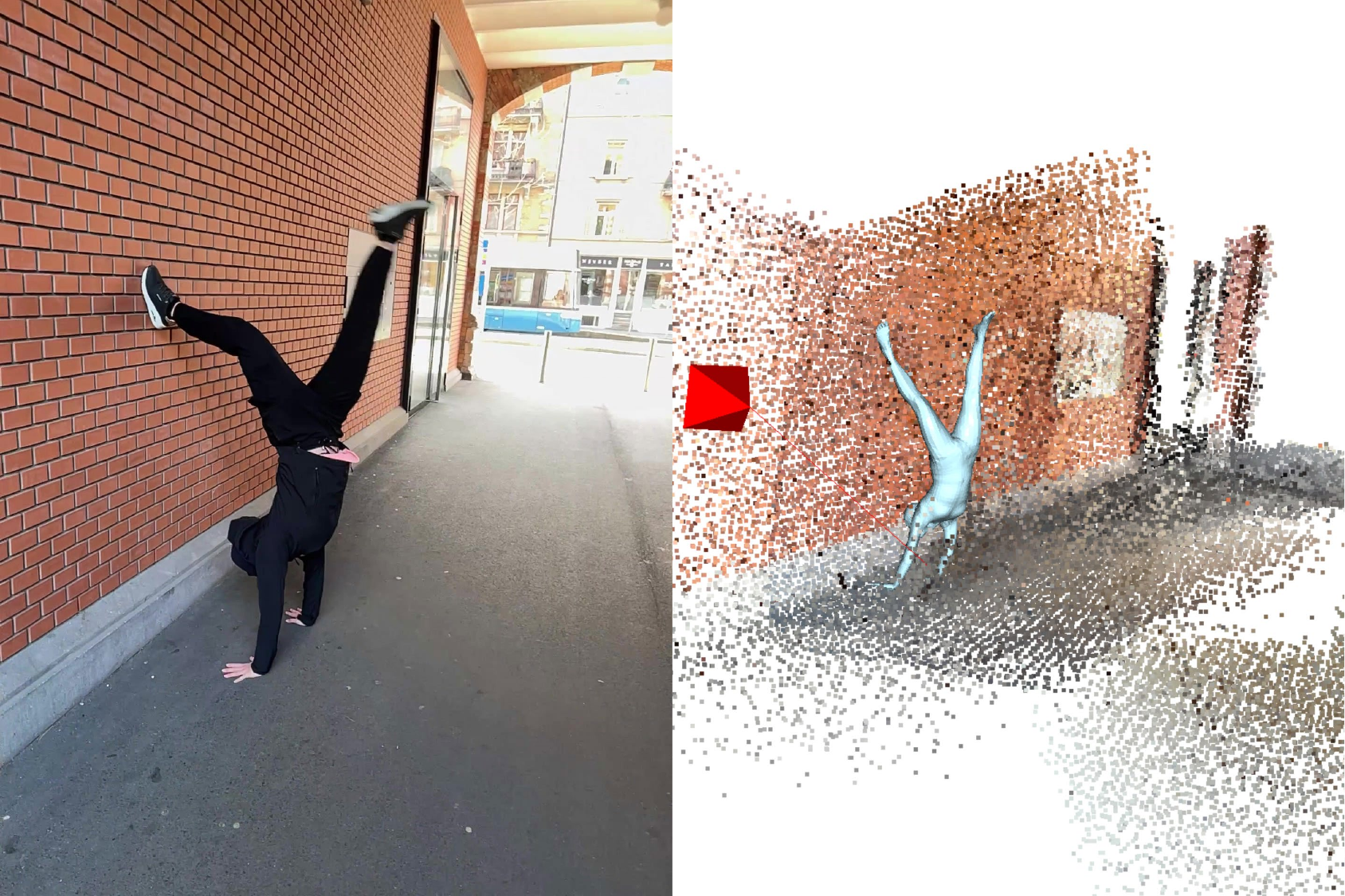}
    \vspace{-0.6cm}
	\caption{Adaptive contact joint identification for a handstand example, where our method correctly identifies the hands as the contact points, ensuring robust scale calibration for unusual postures where the feet are not the real contacts.}
	\label{fig:handstand}
    \vspace{-0.3cm}
\end{figure}

In this section, we provide additional discussion about our method as follows:

\textbf{1) Any module that explicitly detects contacts?} \\
We are not explicitly detecting the human-scene contact joint. Instead, we assume that the lowest human joint in the camera coordinate system's y-axis is the contact joint, as mentioned in Sec.~\textcolor[rgb]{.5,.5,1}{4.2} of the main paper. This strategy identifies feet as the contact points for most cases, given their usual position as the lowest joints. However, in cases where an unusual posture (\eg, a handstand as shown in Fig.~\ref{fig:handstand}) makes another joint the lowest point (\eg, the hands), our method can adaptively identify that joint as the contact point. To validate this assumption, we generated ground-truth contact labels similar to HumanML3D~\cite{humanml3d_2022}, and our strategy achieved an accuracy of 69\% on the EMDB2 dataset. This accuracy level is sufficient for our purposes, as our calibration only requires the joint to be close enough to the scene surface. Additionally, we utilize the median of scale estimates across the entire sequence which further ensures stability and robustness. 

\textbf{2) Why HAC is better than TRAM~\cite{TRAM_2024}?} \\
The key difference between our method and TRAM lies in the source used to calibrate the unknown scale. TRAM relies on background depth predictions from off-the-shelf depth predictors. The depth can be noisy and unreliable even using the SOTA predictor Zoedepth~\cite{Zoedepth_2023}, particularly for scenes with large camera movements and limited background information. In contrast, HAC uses the foreground human—typically the focal point of human-centric videos—as a more stable calibration source. The advantage is extensively discussed and validated in Sec.~\ref{sec:supp:validity}.

\textbf{3) Why not use the fitted ground plane for all cases?} \\
Using one fitted ground plane as the contact surface is not always feasible, particularly in outdoor scenarios where the terrain can vary significantly, such as stairs or uneven surfaces. Therefore, when human-scene contacts are visible and the scene is well reconstructed, it is more accurate to use the reconstructed scene as the contact surface. The fitted ground plane is utilized as a fallback option for scenarios where contact points are out of view, ensuring robust scale calibration under varying conditions.

\end{document}